\documentclass[letterpaper]{article} 
\usepackage{aaai2026}  
\usepackage{times}  
\usepackage{helvet}  
\usepackage{courier}  
\usepackage[hyphens]{url}  
\usepackage{graphicx} 
\usepackage{natbib}  
\usepackage{caption} 
\usepackage{algorithm}
\usepackage{algorithmic}

\usepackage{newfloat}
\usepackage{listings}
\usepackage{bm}
\usepackage{booktabs}
\usepackage{multirow}
\usepackage{amsmath}
\usepackage[table]{xcolor}
\usepackage{cleveref}
\usepackage{amssymb}
\usepackage{xcolor}
\usepackage{comment}
\DeclareCaptionStyle{ruled}{labelfont=normalfont,labelsep=colon,strut=off} 
\floatstyle{ruled}
\newfloat{listing}{tb}{lst}{}
\floatname{listing}{Listing}
\title{LORE: Latent Optimization for Precise Semantic Control \\ in Rectified Flow-based Image Editing}
\author{
    Liangyang Ouyang\textsuperscript{\rm 1}, Jiafeng Mao\textsuperscript{\rm 2}\\
}
\affiliations{
    \textsuperscript{\rm 1}The University of Tokyo, \textsuperscript{\rm 2} CyberAgent\\
    oyly@iis.u-tokyo.ac.jp~~~~~~jiafeng\_mao@cyberagent.co.jp
}

\usepackage{bibentry}

\begin{document}
\nocopyright
\maketitle

\begin{figure*}[h]
    \centering
    \includegraphics[width=\linewidth]{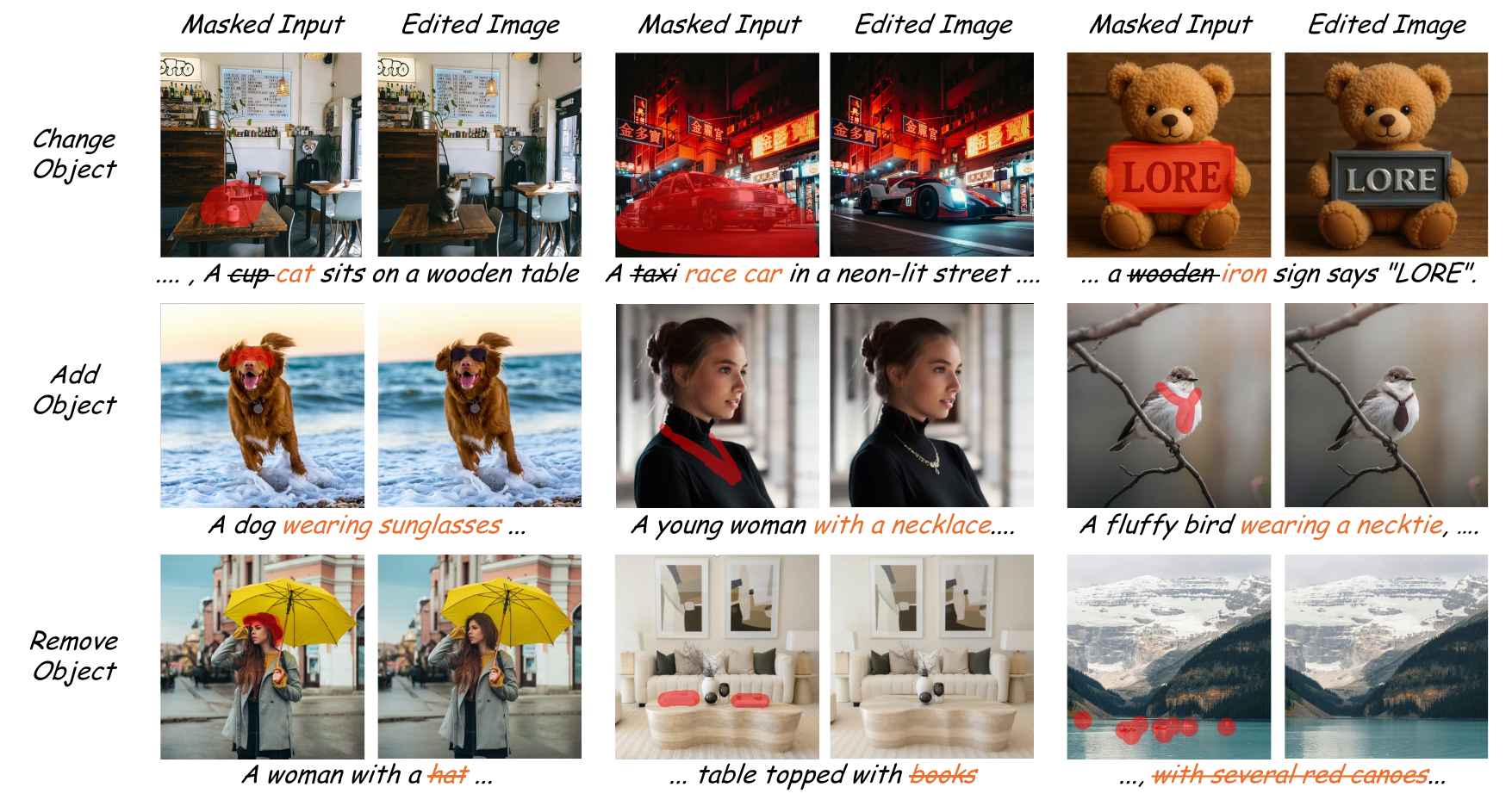}
    \vspace{-20pt}
    \caption{We propose \textit{LORE}, a training-free framework for image editing via \textit{latent optimization} in rectified flow-based diffusion models. \textit{LORE} enables high-quality and controllable edits across a wide range of tasks, including object replacement, addition, removal, and fine-grained attribute modifications. By optimizing the initial noise for a few steps at inference phase, \textit{LORE} achieves efficient editing and accurate background preservation without any model fine-tuning.}
    \label{fig:first}
\end{figure*}

\begin{abstract}
Text-driven image editing enables users to flexibly modify visual content through natural language instructions, and is widely applied to tasks such as semantic object replacement, insertion, and removal. While recent inversion-based editing methods using rectified flow models have achieved promising results in image quality, we identify a structural limitation in their editing behavior: the semantic bias toward the source concept encoded in the inverted noise tends to suppress attention to the target concept. This issue becomes particularly critical when the source and target semantics are dissimilar, where the attention mechanism inherently leads to editing failure or unintended modifications in non-target regions. In this paper, we systematically analyze and validate this structural flaw, and introduce \textit{LORE}, a training-free and efficient image editing method. \textit{LORE} directly optimizes the inverted noise, addressing the core limitations in generalization and controllability of existing approaches, enabling stable, controllable, and general-purpose concept replacement, without requiring architectural modification or model fine-tuning. We conduct comprehensive evaluations on three challenging benchmarks: PIEBench, SmartEdit, and GapEdit. Experimental results show that LORE significantly outperforms strong baselines in terms of semantic alignment, image quality, and background fidelity, demonstrating the effectiveness and scalability of latent-space optimization for general-purpose image editing. Our implementation is available at https://github.com/oyly16/LORE.
\end{abstract}


\section{Introduction}

Text-driven image editing has become a central application in generative modeling, enabling users to modify visual content through natural language instructions. Formally, the task involves an input image and an auxiliary condition—typically a textual prompt or region-specific instruction—specifying the desired change. The goal is to generate an output image in which only the intended regions are modified, with all other content remaining consistent. Among various editing operations, the most fundamental are semantic replacement, insertion, and removal of objects or concepts. These define the core challenge of general-purpose image editing, which is the focus of this work.

Early progress in this area was driven largely by Stable Diffusion \cite{rombach2022high}, whose flexibility and visual quality supported a range of editing pipelines via prompt-based control, mask conditioning, or image inversion \cite{hertzprompt,mokady2023null}. More recently, Rectified Flow (RF)-based diffusion models \cite{liu2023flow}, particularly those incorporating DiT-style transformer backbones \cite{peebles2023scalable}, have significantly improved the fidelity, realism, and semantic alignment of generated images. While these models offer a powerful foundation for editing, their potential remains largely underexplored. Existing RF-based methods are few, and often effective only when the source and target semantics are closely aligned. In contrast, general-purpose editing—where arbitrary objects or concepts must be replaced, inserted, or removed—remains a substantial challenge.

To better understand this limitation, we analyze the editing behavior of RF-based models \cite{flux2024} and identify a systematic bias toward the source concept. As we detailed in experiments, attention maps show that the model tends to focus heavily on the original content, especially when the target semantics differ significantly. This tendency bias suppresses attention signals for the intended concept, leading to failed or low-quality generation. Moreover, the imbalance can cause attention leakage into unrelated regions, unintentionally altering areas that should remain unchanged. Experiments on PIEBench \cite{jupnp} confirm that this issue originates from the semantic prior encoded in the initial noise, which is overlooked by existing methods \cite{wang2024taming,zhu2025kv,avrahami2025stable}.

In this paper, we introduce \textit{LORE}, which stands for \underline{\textit{L}}atent \underline{\textit{O}}ptimization for \underline{\textit{R}}ectified-flow-based \underline{\textit{E}}diting, supporting stable concept replacement across a wide range of semantic variations, while maintaining high-fidelity background consistency by optimizing the initial inverted latent noise. \textit{LORE} consists of two complementary components designed to address the aforementioned challenges. First, an attention-based tendency loss encourages stronger focus on the target concept within the editing region, promoting faithful generation. Second, a masked value injection mechanism preserves features outside the editing mask, ensuring structural and stylistic consistency. These components together enable precise and controllable image editing without any architectural changes or model finetuning.

We evaluate our proposal on three challenging benchmarks, including PIEBench \cite{jupnp}, SmartEdit \cite{huang2024smartedit}, and GapEdit. Results show consistent improvements in semantic alignment, image quality, and background preservation across diverse object categories and editing styles. These findings highlight latent optimization as an effective and scalable solution for general-purpose image editing.

Our main contributions are summarized as follows:
\begin{itemize}
\item We identify a key limitation of current RF-based diffusion models in general-purpose image editing: a strong semantic bias toward the source concept encoded in the initial noise, which suppresses the generation of the intended target.

\item We propose \textit{LORE}, a simple yet effective method for general-purpose image editing that enables accurate semantic replacement and high-fidelity preservation of unrelated regions, without requiring model finetuning.

\item We conduct extensive evaluations on PIEBench, SmartEdit, and GapEdit, demonstrating that \textit{LORE} achieves state-of-the-art performance in semantic alignment, image quality, and background consistency compared to strong baselines.

\end{itemize}

\section{Related Works}

\subsection{Image Editing}
With the advancement of image generation models, image editing methods have also made significant progress \cite{hertzprompt,kawar2023imagic,zhang2023magicbrush,cao2023masactrl,meng2021sdedit,mokady2023null,xia2025dreamomni}. 
These editing methods aim to modify the target object while maintaining background consistency. 
Some approaches fine-tune diffusion models for editing \cite{brooks2023instructpix2pix,geng2024instructdiffusion,lin2024pixwizard,mao2025ace++,Sheynin_2024_CVPR,hou2024high}, but often suffer from limited performance due to the lack of high-quality editing datasets \cite{liu2024referring,huang2024smartedit}.
To enable precisely localized edits, other methods incorporate object masks as input \cite{yang2023dynamic,couairon2023diffedit,tang2024locinv,zhu2025kv,li2024brushedit,mao2025ace++,jang2024identity}.
Our work is the first to optimize the initial noise in a rectified flow-based diffusion model with mask guidance, achieving precise edits without model fine-tuning.

\subsection{RF-based Diffusion Models} 
Recent advances in rectified flow (RF)-based diffusion models \cite{liu2023flow,geng2025mean,esser2024scaling} have introduced a principled reformulation of the denoising process as a continuous normalizing flow, offering improved theoretical foundations and sampling efficiency compared to traditional methods like DDIM \cite{songdenoising}. 
Motivated by their strong generation capabilities, recent works have extended RF-based models to image editing tasks via inversion and conditioning strategies \cite{avrahami2025stable,chen2024training,feng2025dit4edit,kulikov2024flowedit,wang2024taming,zhu2025kv,xu2025unveil,Dalva_2025_CVPR,Zhu_2024_CVPR,deng2024fireflow,jiao2025uniedit,rout2024semantic,tan2024ominicontrol}. Among them, RFEdit \cite{wang2024taming} introduces a quadratic reconstruction term to enhance fidelity during inversion. StableFlow \cite{avrahami2025stable} identifies vital layers within DiT-based RF models, editing images by manipulating intermediate features. KVEdit \cite{zhu2025kv} incorporates key-value pairs with spatial masks to improve attention-driven modifications.
Building on these advances, our method \textit{LORE} is the first to perform latent optimization within an RF-based diffusion framework, enabling fine-grained, region-aware semantic editing.

\subsection{Latent Optimization} 
Initial noise serves as the starting point of the generation process in diffusion models \cite{liu2023flow,ho2020denoising}. 
Although it is sampled from a standard Gaussian distribution, different noise samples can exhibit distinct generation tendencies. 
Prior studies have shown that manipulating the initial noise can control various aspects of the output, such as object semantic \cite{mao2023guided,chefer2023attend,yuanfreqprior}, sketch \cite{ding2024training}, quality \cite{qi2024not,xu2025good}, style \cite{cui2024instastyle} and layout \cite{kikuchi2021constrained,mao2024lottery,Shirakawa_2024_CVPR,sun2024spatial}.
Building on these works, we investigate the generation bias of initial noise in RF-based editing methods, and propose a novel latent optimization approach to guide the editing process more precisely.

\begin{figure*}[t]
    \centering
    \includegraphics[width=0.95\linewidth]{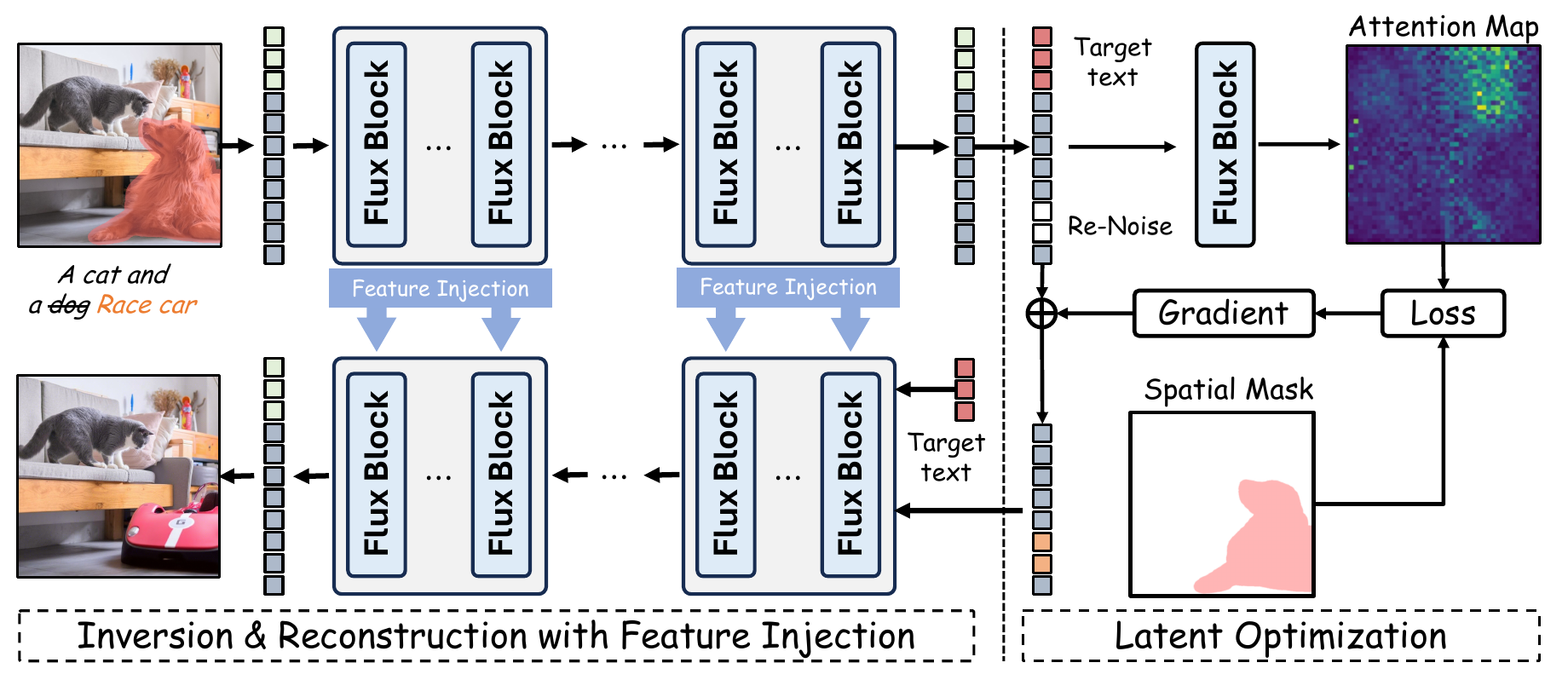}
    \caption{Framework of \textit{LORE}. Our method consists of three steps: (1) inversion to get inverted noise, (2) optimizing the inverted noise according to attention map and (3) reconstruction with value injection.}
    \vspace{-10pt}
    \label{fig:method}
\end{figure*}

\section{Approach}

\subsection{Preliminary}
\subsubsection{Inversion in Rectified Flow Models}
Rectified Flow (RF) models \cite{liu2023flow} construct a straight path between the noise distribution and the real data distribution by learning a time-dependent velocity field $v_\theta(\bm{x}, t)$. This process can be formulated as an ordinary differential equation (ODE):
\begin{equation}
    d\bm{x}_t = v_\theta(\bm{x}, t)dt, t \in [0, 1]
\end{equation}
Due to the reversible nature of this ODE, RF-based models support image editing through a two-stage process: inversion and denoising.  
Given a source image $\mathcal{I}_\text{source}$, inversion maps it to a latent noise $z_0$ by integrating the ODE backward from $t=1$ to $t=0$ under the source prompt. This latent $z_0$ is then edited via forward integration with a target prompt to obtain the modified image $\mathcal{I}_\text{target}$. Importantly, the inverted noise  $z_0$ can be viewed as a \textit{semantic embedding} of the image and carries meaningful information about the image structure, content, and appearance, which makes it a suitable medium for the image editing task.

\subsubsection{Generation Tendency of Inverted Latent} 

In diffusion models, attention maps calculated by word tokens and visual tokens indicate where and how the semantics of each prompt word are grounded in the generated image. Specifically, the attention map $M^{(i)}$ corresponding to a prompt token $p^{(i)}$ highlights the spatial regions in the image that are most semantically aligned with that token. Recent studies have shown that the layout and shape of generated objects are highly correlated with these attention maps \cite{hertzprompt,cao2023masactrl}. This is because the attention operation distributes features from the value matrix $V$, which contains rich semantic information for each word. In essence, cross-attention computes the similarity between each image patch in the latent $z_t$ and each prompt token $p^{(i)}$, and reinforces features of $p^{(i)}$ in regions that are already semantically similar. This creates a positive feedback loop that amplifies semantic alignment, which is referred to as the \textit{generation tendency}.

However, in inversion-based image editing, we observe that the inverted noise inherently retains a strong bias toward reconstructing the original image. This property is what makes inversion suitable for editing tasks in the first place, and it poses no issue when only minor or semantically similar modifications are required. However, when the editing involves replacing a source concept with a semantically dissimilar target concept in the same location, this generation bias becomes problematic. The region corresponding to the source object exhibits high similarity with its original concept and, consequently, low similarity with the new target concept. This suppresses attention activation for the target token in the intended region, impeding the correct allocation of semantic features. As a result, the model may fail to generate the target object, degrade image quality, or even alter regions that should remain unchanged.
A dedicated analysis is conducted to quantify generation tendencies of source and target concepts under different types of initial noise latent in the section Experiments.

\subsection{\textit{LORE}}
Based on our earlier analysis, we identify that the core failure in existing image editing models—especially in general object replacement tasks—lies in inaccurate attention responses: the target region often lacks sufficient activation for the desired concept, while unrelated regions exhibit undesired activations.  
To address these two issues, \textit{LORE} consists of two main components. First, we optimize the inverted noise to enhance the target region’s responsiveness to the desired concept, thereby providing a better initialization for the denoising process. Second, we introduce Masked Value Injection to explicitly preserve the features in unedited regions, improving background consistency and preventing semantic leakage.

\subsubsection{Latent Optimization via Tendency Loss}

A key design choice in our framework is to optimize the inverted noise, i.e., the input to the first denoising step. Prior studies have demonstrated that in deterministic denoising processes, the initial noise plays a decisive role in shaping the generative trajectory. Owing to the positive feedback loop between the latent features and attention-based allocation, modifying only the initial noise is sufficient to steer the entire generation process.
Therefore, we apply gradient-based optimization to the inverted noise $z_0$, which incurs minimal computational overhead and avoids interfering with the model architecture or denoising dynamics.

Our objective is to enhance the attention response to the target concept within the editing region. Intuitively, the attention map corresponding to the target token should exhibit high and spatially focused activation within the masked area. To achieve this, we apply a Gaussian smoothing operator $G(\cdot)$ to the attention map and maximize its peak value inside the mask. Specifically, for an attention layer with parameters $\mathcal{W}$, the attention output $\mathcal{A}$ is computed as:

\begin{equation}
\mathcal{A}(Q, K) = \text{softmax}\left(\frac{QK^\top}{\sqrt{d}}\right),
\label{eq:attention}
\end{equation}
where $Q = \mathcal{W}_Q \cdot z_0$ and $K = \mathcal{W}_K \cdot \mathcal{P}$.
We extract the slice $\mathcal{A}_{\text{obj}} \in \mathcal{A}$ corresponding to the target object token and our optimization objective can be formulated as follows,

\begin{equation}
\arg\min_{z_0} \mathcal{L} = 1 - \max\left( \mathcal{M} \cdot G(\mathcal{A}_{\text{obj}}) \right),
\end{equation}
where $\mathcal{M}$ denotes the spatial editing mask.
This objective encourages strong, focused attention within the target region while avoiding uniformly high or unnatural activations that may degrade generation quality. 

\subsubsection{Masked Value Injection}

Feature injection into Transformer layers has been widely adopted in various tasks, such as key-value (KV) caching in language models. In the context of image editing, some prior works attempt to preserve fidelity by injecting the key and value features from inversion phase into the denoising process \cite{avrahami2025stable, zhu2025kv}. While this strategy can be effective in some scenarios, it suffers from structural limitations. Specifically, $V$ encodes the rich visual features of each token, while $K$ controls the retrieval mechanism that guides value aggregation. Injecting both $K$ and $V$ from the original image into the denoising phase can easily result in semantic mismatch, as the injected features may not align with the edited prompt. Even re-computing $K$ and $V$ in the target region by resampling noise often fails to recover coherent spatial structure.

\textit{LORE} takes a fundamentally different approach. Since our optimized latent already encodes a generation tendency aligned with the target semantics, we do not intervene in the key representations at all. Instead, we simply reuse the $V$ values from the non-edited regions recorded during inversion, and inject them back during denoising as follows,

\begin{equation}
\hat{v}_t \leftarrow (1 - \mathcal{M}) \cdot v_t + \mathcal{M} \cdot \hat{v}_t
\end{equation}
where $v_t$ and $\hat{v}_t$ denote value calculated during inversion and denoising, respectively. This simple yet effective strategy preserves background content without interfering with the semantics in the edited region.

\begin{table*}[t]
\centering
\scriptsize
\setlength{\tabcolsep}{1.5 mm}
\begin{tabular}{l|cccccc|cccccc|cccccc}
\toprule
\multirow{3}{*}{Method} & \multicolumn{6}{c|}{PIEBench} & \multicolumn{6}{c|}{SmartEdit} & \multicolumn{6}{c}{GapEdit}\\
\cmidrule(lr){2-19}
& \multicolumn{2}{c}{Text Align$\uparrow$} & \multicolumn{2}{c}{Image Quality$\uparrow$} & \multicolumn{2}{c|}{Consistency$\downarrow$} & \multicolumn{2}{c}{Text Align$\uparrow$} & \multicolumn{2}{c}{Image Quality$\uparrow$} & \multicolumn{2}{c|}{Consistency$\downarrow$} & \multicolumn{2}{c}{Text Align$\uparrow$} & \multicolumn{2}{c}{Image Quality$\uparrow$} & \multicolumn{2}{c}{Consistency$\downarrow$} \\
\cmidrule(lr){2-19}
 & CLIP & IR & HPS & AS & LPIPS & MSE & CLIP & IR & HPS & AS & LPIPS & MSE & CLIP &IR & HPS & AS & LPIPS & MSE \\
\midrule
VAE & $\textcolor{gray}{23.4}$ & $\textcolor{gray}{-51.6}$ & $\textcolor{gray}{25.1}$ & $\textcolor{gray}{5.82}$ & $\textcolor{gray}{3.8}$ & $\textcolor{gray}{2.1}$ & $\textcolor{gray}{22.0}$ & $\textcolor{gray}{-93.5}$ & $\textcolor{gray}{25.5}$ & $\textcolor{gray}{5.44}$ & $\textcolor{gray}{7.6}$ & $\textcolor{gray}{17.5}$ & $\textcolor{gray}{20.0}$ & $\textcolor{gray}{-141}$ & $\textcolor{gray}{22.5}$ & $\textcolor{gray}{5.61}$ & $\textcolor{gray}{4.7}$ & $\textcolor{gray}{6.5}$\\
\midrule
\rowcolor{gray!20}
RF-Edit & $24.8$ & $3.0$ & $26.8$ & $5.91$ & $18.0$ & $32.0$ & $23.9$ & $-39.4$ & $27.2$ & $\bm{5.61}$ & $19.5$ & $23.8$
& $21.0$ & $-116$ & $23.7$ & $\bm{5.70}$ & $19.6$ & $14.7$
\\
\rowcolor{gray!20}
StableFlow & $24.8$ & $16.3$ & $26.2$ & $5.58$ & $16.1$ & $35.3$ & $23.6$ & $-22.9$ & $26.0$ & $5.29$ & $19.1$ & $40.2$
& $22.8$ & $-70.6$ & $23.1$ & $5.24$ & $19.6$ & $39.4$
\\
\rowcolor{gray!20}
FlowEdit & $26.2$ & $70.2$ & $27.0$ & $5.82$ & $24.5$ & $40.2$ & $24.6$ & $17.6$ & $26.4$ & $5.43$ & $13.2$ & $44.3$
& $24.8$ & $4.2$ & $25.6$ & $5.62$ & $25.7$ & $39.8$
\\
\midrule
KV-Edit & $25.6$ & $51.0$ & $26.5$ & $5.76$ & $10.9$ & $22.2$ & $23.4$ & $-45.8$ & $25.6$ & $5.35$ & $9.9$ & $17.7$
& $23.6$ & $-42.6$ & $23.6$ & $5.40$ & $9.3$ & $7.9$
\\
FLUX.Fill & $24.0$ & $-0.6$ & $23.8$ & $5.55$ & $19.1$ & $26.4$
& $24.5$ & $3.5$ & $26.1$ & $5.36$ & $10.2$ & $18.0$
& $23.2$ & $-51.2$ & $23.3$ & $5.36$ & $9.9$ & $7.4$
\\
ACE++ & $24.7$ & $31.4$ & $25.5$ & $5.70$ & $13.9$ & $28.6$ & $24.8$ & $0.1$ & $26.4$ & $5.42$ & $10.1$ & $17.6$ & $23.7$ & $-42.7$ & $23.8$ & $5.37$ & $9.3$ & $8.5$

\\
\midrule

$\textit{LORE}_\text{(Ours)}$ & $\bm{26.3}$ & $\bm{75.7}$ & $\bm{27.5}$ & $\bm{5.94}$ & $\bm{10.8}$ & $\bm{18.1}$ & $\bm{25.4}$ & $\bm{21.7}$ & $\bm{26.9}$ & $5.45$ & $\bm{9.8}$ & $\bm{17.7}$ & $\bm{26.3}$ & $\bm{31.2}$ & $\bm{25.6}$ & $5.54$ & $\bm{9.3}$ & $\bm{7.4}$

\\

\bottomrule
\end{tabular}
\caption{Editing Results on PIEBench, SmartEdit and  GapEdit. Rows with a gray background indicate methods that do not accept mask inputs.}
\label{tab:main}
\end{table*}

\section{Experiments}
\label{sec:experiments}
\subsection{Evaluation Settings}
\noindent\textbf{Datasets}
We conduct experiments on three datasets that collectively cover a broad range of general image editing scenarios, including typical edits, instance-level edits, and challenging semantic-shift cases.

\begin{itemize}
\item \textbf{PIE-Bench}\cite{jupnp} is a widely used image editing benchmark. We removed examples that do not involve a target object, such as object deletion or style change tasks, resulting in a final set of 484 samples.

\item \textbf{SmartEdit}\cite{huang2024smartedit} focuses on editing one object instance among multiple similar instances. We use 131 samples with clearly defined object categories.

\item \textbf{GapEdit} is a benchmark we construct to evaluate editing performance under large semantic gaps between source and target objects. We first select 20 frequently occurring source objects from PIEBench and SmartEdit, then identify semantically distant target categories using FLUX’s T5 encoder \cite{raffel2020exploring}. Representative source images are chosen from the two datasets, resulting in 174 samples forming a challenging dataset for semantic-gap evaluation. 

\end{itemize}

\noindent\textbf{Baseline Models}
We select six recent RF-based image editing methods as baselines, covering both mask-free and mask-based approaches. Since all these methods are built upon FLUX framework, we additionally include FLUX VAE reconstruction results in \cref{tab:main} as a reference. The selected baselines are as follows,
\begin{itemize}
\item \textbf{RF-Edit} \cite{wang2024taming} improves editing quality by introducing a second-order term during inversion and applying value injection in early denoising steps. 

\item \textbf{StableFlow} \cite{avrahami2025stable} utilizes vital layers in FLUX to facilitate stable image editing. Its ability to edit real images is limited, as the method is primarily designed for editing generated images.

\item \textbf{FlowEdit} \cite{kulikov2024flowedit} constructs an ODE that directly maps between the source and target distributions, enabling inversion-free image editing.

\item \textbf{KV-Edit} \cite{zhu2025kv} builds on FLUX by incorporating masked key and value replacement along with random renoising, achieving precise editing results. 

\item \textbf{FLUX.Fill} \cite{flux2024} is a RF transformer model designed for image inpainting conditioned on text prompts. It accepts mask inputs for filling specified regions.

\item \textbf{ACE++} \cite{mao2025ace++} builds upon the FLUX model by incorporating LoRA fine-tuning, enabling better generalization across diverse datasets and editing tasks. It also supports mask-based inputs, allowing it to perform localized image editing effectively.
\end{itemize}

\noindent\textbf{Metrics}
Following previous works \cite{zhu2025kv}, we adopt six image editing evaluation metrics. For text alignment, we use CLIP similarity \cite{radford2021learning} and Image Reward\cite{xu2023imagereward}. For image quality, we employ HPS \cite{zhang2018unreasonable} and Aesthetic Score (AS) \cite{schuhmann2022laion}. For background consistency, we measure LPIPS \cite{zhang2018unreasonable} and MSE$_{\times 10^3}$ between the edited and source images in the unedited regions (outside the mask).

\noindent\textbf{Implementation Details}
All our methods and baselines are implemented based on the FLUX codebase \cite{flux2024}. Experiments are conducted on a single NVIDIA A100 GPU. For latent optimization, we use a simple SGD optimizer with a learning rate of 0.01, running for 10 iterations. For fair comparison, the denoising step is set to 15 and guidance scale is set as 2 for all methods. We disable re-noising in inversion-based methods \textit{LORE, KVEdit, and RFEdit}. 

\subsection{Editing Results}

\begin{figure*}[t]
    \centering
    \includegraphics[width=\linewidth]{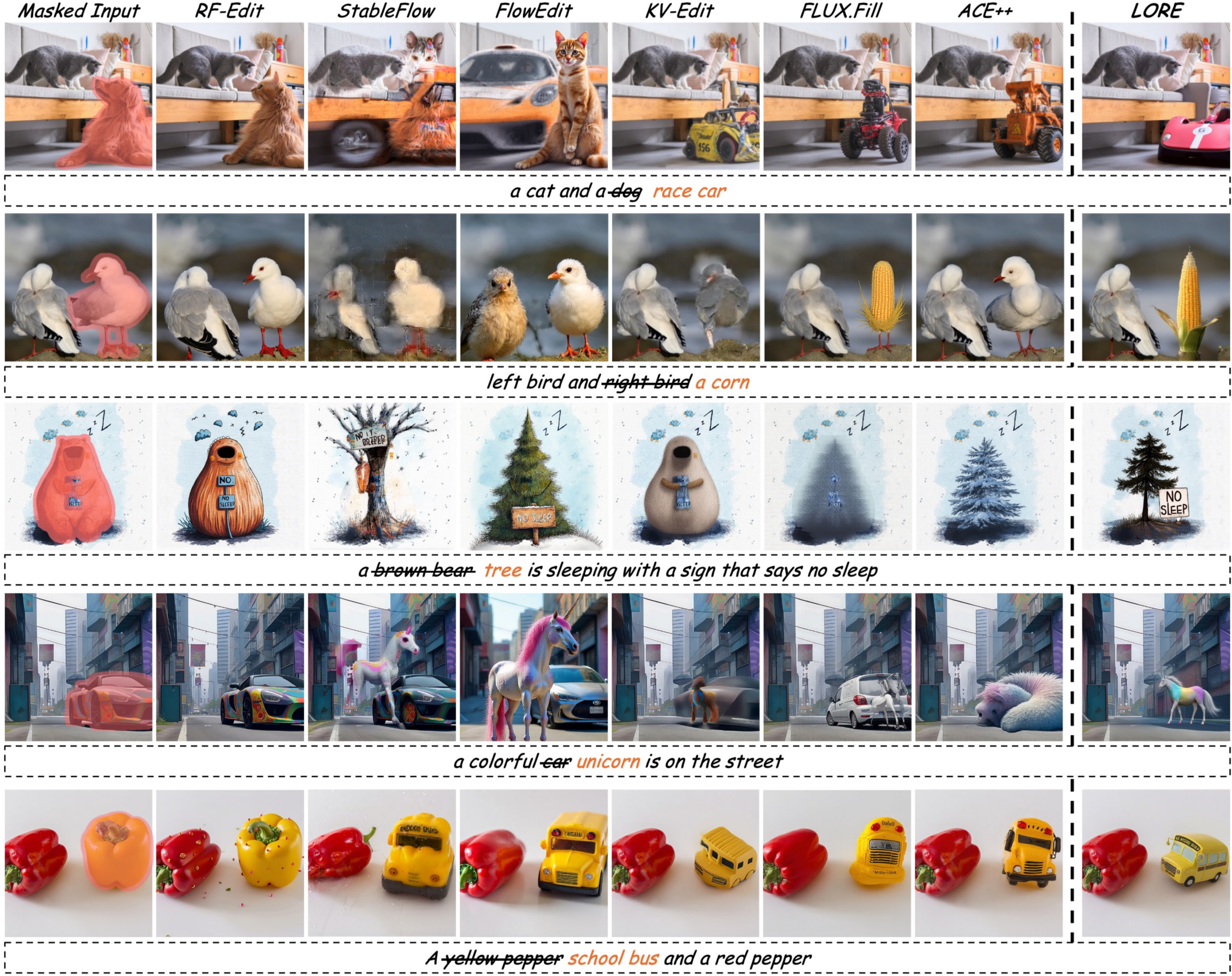}
    \caption{Editing results comparing with other methods.}
    \label{fig:result}
\end{figure*}

\noindent\textbf{Quantitative Comparison}
We present the quantitative results across the three datasets in \cref{tab:main}.
Our method consistently outperforms all baselines in text alignment metrics, with especially notable improvements on SmartEdit and GapEdit, demonstrating the effectiveness of our latent optimization strategy in guiding generation tendency.
For image quality, our approach surpasses most baselines, while RFEdit shows slightly higher scores in the AS metric, but RFEdit frequently fails to replace objects in SmartEdit and GapEdit.
In terms of background consistency, our method remains on par with other methods that utilize mask-based inputs, and significantly better than baselines without mask guidance.
Overall, metric results confirm that our method achieves state-of-the-art editing performance across all three datasets.

\noindent\textbf{Qualitative Comparison} As illustrated in \cref{fig:result}, \textit{LORE} shows stable and successful editing results. 
Methods without mask inputs, such as RF-Edit, StableFlow, and FlowEdit, often struggle in complex scenes where multiple objects co-exists. In contrast, approaches that take mask inputs, such as KV-Edit, FLUXFill and ACE++, can focus on the intended editing area. However, due to the mismatch between the generation tendency in the inversion noise and the semantics of the target object, KV-Edit often produces unstable results. FLUXFill and ACE++ are specifically fine-tuned for masked image editing tasks. They generate images by denoising from random noise, which can also be inconsistent with the target prompt. With optimizing the inverted noise to align with the target prompt, the editing of \textit{LORE} becomes more stable and semantically accurate. The examples clearly demonstrate that the effectiveness of our method becomes significant in complex scenes and when there is a large semantic gap between the source and target objects.

\subsection{Ablation Study}

\subsubsection{Learning Rate} 
\begin{figure}[t]
    \centering
    \includegraphics[width=\linewidth]{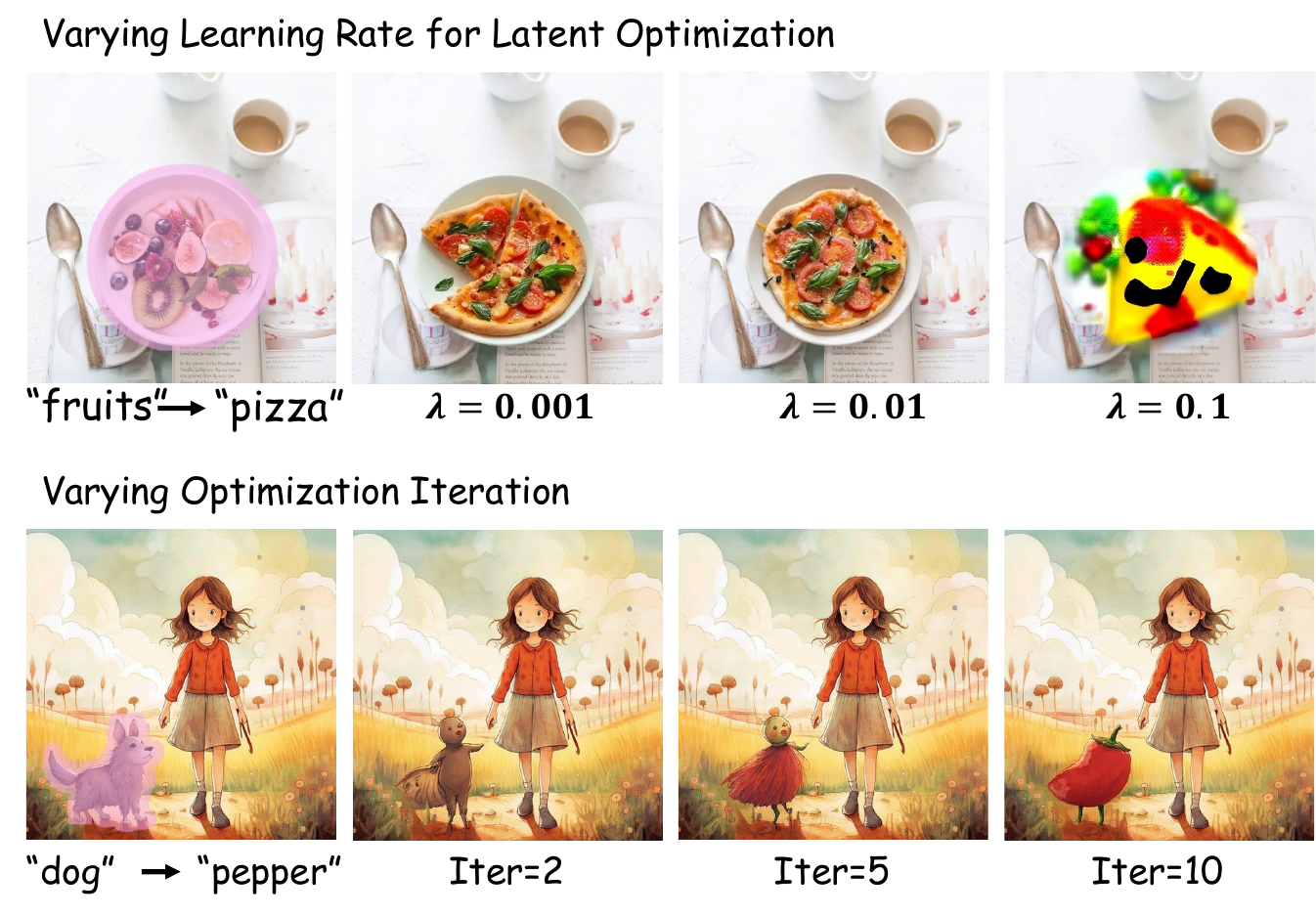}
    \vspace{-20pt}
    \caption{Quality analysis on latent optimization learning rate and iteration numbers.}
    \vspace{-10pt}
    \label{fig:ablation}
\end{figure}

In \cref{tab:lr}, we compare the impact of different learning rates on GapEdit. A large learning rate may disrupt the structure of the noise distribution, leading to unstable outputs, while a small learning rate fails to effectively optimize the noise and cannot guide the generation tendency. 
As shown in \cref{fig:ablation}, a learning rate of $0.001$ results in an incomplete pizza, while $0.1$ leads to a corrupted image due to over-updating of the noise.

\begin{table}[t]
\centering
\setlength{\tabcolsep}{2mm}
\begin{tabular}{l|cc|cc|cc}
\toprule
\multirow{2}{*}{$\lambda$} & \multicolumn{2}{c}{Align$\uparrow$} & \multicolumn{2}{c}{Quality$\uparrow$} & \multicolumn{2}{c}{Consistency$\downarrow$} \\
\cmidrule(lr){2-7} 
 & CLIP & IR & HPS & AS & LPIPS & MSE \\
\midrule
$0$ & $25.5$ & $9.7$ & $25.3$ & $5.51$ & $\bm{8.7}$ & $\bm{7.1}$ \\
$10^{-1}$ & $22.3$ & $-1.0$ & $20.3$ & $4.49$ & $14.0$ & $11.0$
 \\
$10^{-2}$ & $\bm{26.3}$ & $\bm{31.2}$ & $\bm{25.6}$ & $\bm{5.54}$ & $9.3$ & $7.4$\\
$10^{-3}$ & $\underline{25.7}$ & $\underline{13.0}$ & $\underline{25.4}$ & $\underline{5.52}$ & $8.8$ & $\bm{7.1}$
\\
$10^{-4}$ & $25.4$ & $8.4$ & $25.3$ & $5.51$ & $\bm{8.7}$ & $\bm{7.1}$
 \\
\bottomrule
\end{tabular}
\caption{Ablation on learning rate of latent optimization.}
\vspace{-20pt}
\label{tab:lr}
\end{table}

\noindent\textbf{Optimization Iterations} As shown in \cref{tab:epoch}, we conduct an ablation study on GapEdit dataset by varying the number of latent optimization iterations. We observe that certain number of steps lead to optimal results. When the number of iterations is too small, the optimized noise remains close to the inverted noise, failing to introduce sufficient semantic change. On the other hand, excessive optimization tends to disrupt the structural information encoded in the noise, resulting in degraded image quality. We further provide a visualization study that demonstrates the controllability enabled by adjusting the number of latent optimization iterations. As illustrated in \cref{fig:ablation}, varying the iteration effectively controls the semantic distance between the source and target objects in the generated results. 

\begin{table}[t]
\centering
\small
\setlength{\tabcolsep}{1.5mm}
\begin{tabular}{l|cc|cc|cc|c}
\toprule
\multirow{2}{*}{$\mathcal{N}_\text{iter}$} & \multicolumn{2}{c}{Align$\uparrow$} & \multicolumn{2}{c}{Quality$\uparrow$} & \multicolumn{2}{c}{Consistency$\downarrow$} & Time Cost\\
\cmidrule(lr){2-8}
 & CLIP & IR & HPS & AS & LPIPS & MSE & Sec.\\
\midrule
$0$ & $25.5$ & $9.7$ & $25.3$ & $5.51$ & $\bm{8.7}$ & $\bm{7.1}$ & $17.2$\\
$2$ & $25.9$ & $23.5$ & $25.5$ & $5.51$ & $8.8$ & $7.2$
 & $19.3$\\
$5$ & $26.0$ & $25.4$ & $\bm{25.6}$ & $5.52$ & $9.0$ & $7.4$
& $21.5$\\
$10$ & $\bm{26.3}$ & $\bm{31.2}$ & $\bm{25.6}$ & $\bm{5.54}$ & $9.3$ & $7.4$ & $28.0$\\
$20$ & $25.7$ & $16.3$ & $25.2$ & $5.45$ & $9.6$ & $7.6$ & $36.4$\\
\bottomrule
\end{tabular}
\caption{Ablation on iterations of latent optimization.}
\label{tab:epoch}
\end{table}


\noindent\textbf{Time Cost} We analyze the computational cost introduced by \textit{LORE} in \cref{tab:epoch}. Since each optimization step involves a forward pass through the denoising model, training iterations are equivalent to performing multiple additional denoising steps. We recommend using 5–10 optimization iterations, which only increases the generation time by $25\%$ while significantly improving editing stability and visual fidelity.

\subsection{Analysis on Generation Tendency}

To further understand how the initial noise influences the model’s generation behavior, we conduct an analysis focused on generation tendency. Specifically, we investigate whether the model tends to generate the source or target object depending on the type of initial noise used.

We use $420$ examples from the PIEBench dataset \cite{jupnp}, each involving the replacement of a source concept $\mathcal{C_\text{src}}$ with a target concept $\mathcal{C_\text{tgt}}$. For each example, we focus on the first denoising step of the FLUX model and extract the cross-attention maps corresponding to $\mathcal{C_\text{src}}$ and $\mathcal{C_\text{tgt}}$. We then compute the average attention weight within the object mask $\mathcal{M}$ as an indicator of generation tendency:
\begin{equation}
\text{Tend}(c) = \text{mean}(\mathcal{M} \cdot \mathcal{A}[c]), c\in \{\mathcal{C}_\text{src},\mathcal{C}_\text{tgt}\}
\end{equation}
where $\mathcal{A}$ is the attention map calculated as \cref{eq:attention}. We compare generation tendency of $\mathcal{C_\text{src}}$ and $\mathcal{C_\text{tgt}}$ across three types of initial noise: noise $z_0$ inverted from source image, random Gaussian noise $z_\text{random}$, and optimized noise $\hat{z}_0$ produced by our latent optimization.

\begin{table}[t]
\centering
\setlength{\tabcolsep}{5mm}
\begin{tabular}{l|ccc}
\toprule
~ & $z_o$ & $z_\text{random}$ & $\hat{z}_0$  \\
\midrule
$\mathcal{C}_\text{src}~@~\mathcal{P}_\text{src}$ & \textbf{27.3} & 27.0 & 28.7 \\
$\mathcal{C}_\text{tgt}~@~\mathcal{P}_\text{tgt}$ & 25.8 & 26.2 & \textbf{29.0} \\
\midrule
$\mathcal{C}_\text{src}~@~\mathcal{P}_\text{comb}$ & \textbf{34.5} & 33.9 & 33.4 \\
$\mathcal{C}_\text{tgt}~@~\mathcal{P}_\text{comb}$ & 33.3 & 33.6 & \textbf{34.7} \\
\bottomrule
\end{tabular}
\caption{Generation Tendency (\%) calculated by different types of initial noise. $\mathcal{C}_\text{src}$ and $\mathcal{C}_\text{tgt}$ denotes the source concept and target concept respectively. $\mathcal{P}_\text{src}$ and $\mathcal{P}_\text{tgt}$ indicate the source prompt and prompt after words replacement. $\mathcal{P}_\text{comb}$ indicates a specifically created prompts contain both source concept and target concept.}
\label{tab:tendency}
\end{table}

\begin{figure}[t]
    \centering
    \includegraphics[width=\linewidth]{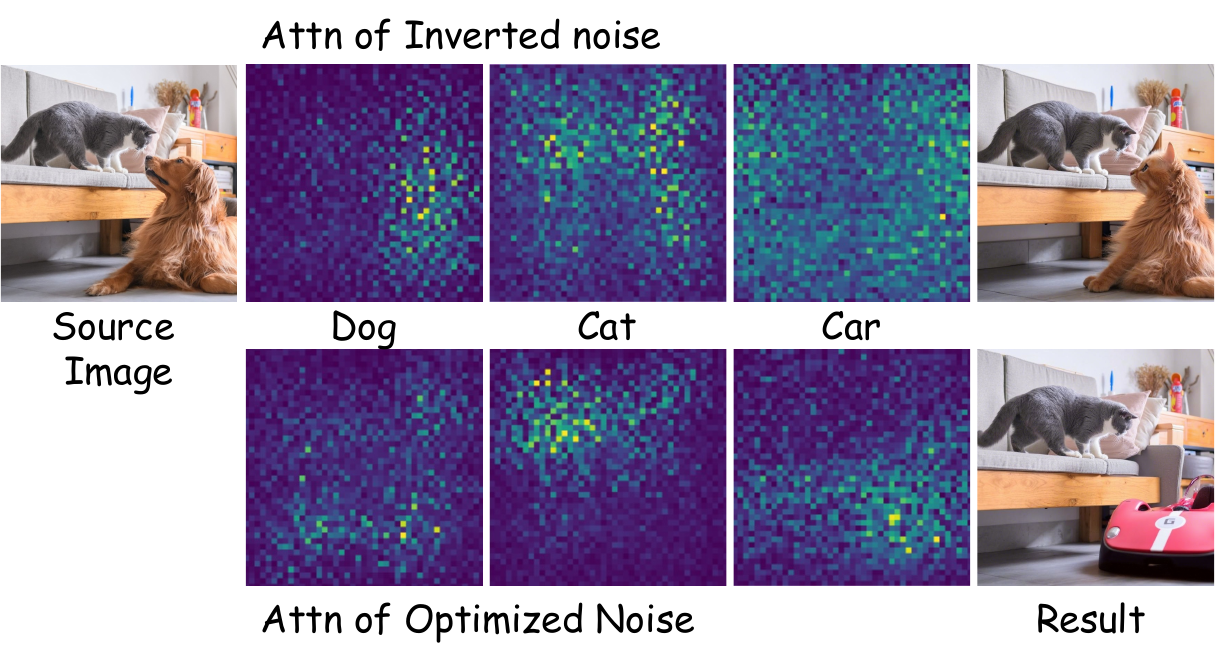}
    \vspace{-20pt}
    \caption{Attention maps of inverted and optimized noise.}

    \vspace{-10pt}
    \label{fig:analyze}
\end{figure}

As shown in \cref{tab:tendency}, both inversion and random noise show a higher attention response for the source object than for the target object, indicating a natural bias toward regenerating the original content. This is expected: inversion noise is obtained under the guidance of the source prompt, and the mask aligns with the shape of the source object. Even with random noise, the source concept may still dominate due to its stronger contextual plausibility. However, when using optimized noise, the attention for the target object increases significantly, surpassing that of the source. This indicates that our latent optimization successfully adjusts the generation tendency to favor the target object. 
\subsubsection{Generation Tendency Visualization}
As shown in \cref{fig:analyze}, for inverted noise, the attention map for \verb|car| is dispersed and weak, leading the model to mistakenly generate a \verb|cat| instead. In contrast, our optimized noise yields a strong, focused attention map for \verb|car| within the correct region, resulting in accurate editing. This supports our core hypothesis: optimizing the latent input can rectify generation tendency and improve editing fidelity without altering the model architecture.

\section{Conclusion}

In this work, we identify a key limitation of rectified flow-based diffusion models for image editing: the generation tendency embedded in the initial inversion noise. To address this issue, we propose \textit{LORE}, a latent optimization framework that refines the inversion noise via an attention-based loss and masked value injection. \textit{LORE} enables precise and robust semantic edits while preserving background consistency. Extensive experiments on PIE-Bench, SmartEdit, and GapEdit demonstrate that our method achieves superior performance over existing approaches in semantic alignment, visual quality, and consistency.

\bibliography{aaai2026}
\end{document}